\documentclass[11pt]{article}

\usepackage[preprint]{acl}

\usepackage{times}
\usepackage{latexsym}

\usepackage[T1]{fontenc}

\usepackage[utf8]{inputenc}

\usepackage{microtype}

\usepackage{inconsolata}

\usepackage{graphicx}

\usepackage{amssymb}
\usepackage{etoolbox}
\usepackage[most]{tcolorbox}
\usepackage{subcaption}
\usepackage{booktabs}

\newtcolorbox{promptbox}[1][]{
  colback=green!7,   
  colframe=black,    
  boxrule=0.8pt,     
  arc=2pt,           
  left=6pt, right=6pt, top=6pt, bottom=6pt,
  #1
}

\title{When Models Decide and When They Bind: A Two-Stage Computation for Multiple-Choice Question-Answering}

\author{Hugh Mee Wong, Rick Nouwen, Albert Gatt \\
        Utrecht University \\
        Utrecht, The Netherlands \\
        \texttt{\{h.m.wong, r.w.f.nouwen, a.gatt\}@uu.nl}
}

\begin{document}
\maketitle
\begin{abstract}
Multiple-choice question answering (MCQA) is easy to evaluate but adds a meta-task: models must both solve the problem and output the symbol that \emph{represents} the answer, conflating reasoning errors with symbol-binding failures. We study how language models implement MCQA internally using representational analyses (PCA, linear probes) as well as causal interventions.
We find that option-boundary (newline) residual states often contain strong linearly decodable signals related to per-option correctness.
Winner-identity probing reveals a two-stage progression: the winning \emph{content position} becomes decodable immediately after the final option is processed, while the \emph{output symbol} is represented closer to the answer emission position.
Tests under symbol and content permutations support a two-stage mechanism in which models first select a winner in content space and then bind or route that winner to the appropriate symbol to emit.
We release our code at \url{https://github.com/to-reveal-after-reviews}.
\end{abstract}

\section{Introduction}
Multiple-choice question answering (MCQA) is pervasive in natural language processing (NLP) benchmarks, in part because of its ease of evaluation compared to long-form generation:
assessing a model's response can often be reduced to inspecting the token (distribution) immediately following the user's prompt.
However, reframing NLP tasks as MCQA prompts also introduces an additional meta-task, over and above the NLP task itself.
Specifically, a language model must both determine the correct answer to the target question and recognize that it should output the symbol that \emph{represents} that answer.
Consequently, when a model selects an incorrect option, this failure may stem from a misunderstanding of the target question, the MCQA format, or both.

Prior work has shown that language models exhibit selection biases for particular option identifiers such as the label ``A'' \citep{zheng2024-selection, dominguez-olmedo2024-survey, li2025-anchor, xue2024-binding}.
They may also be sensitive to the ordering of answer choices \citep{pezeshkpour2024-order}, the specific symbols used to denote the options \citep{yang2025-symbols, wiegreffe2025-mcqa}, and the length of answer options \citep{zhao2025-length}.
Furthermore,
the answer obtained from comparing probabilities of the first tokens in MCQA is often not consistent with the long-form generation output \citep{wang2024-probabilities, tsvilodub2024-robust}.

These issues underscore the need to better understand \emph{how} a language model solves MCQA tasks, separating
a model's task-relevant knowledge from artifacts introduced by the MCQA format itself, especially
\emph{symbol binding}, where a model has to associate an answer symbol with its corresponding option text.
Mechanistic interpretability (MI) methods can be used to identify why and how answer options are represented, as well as how symbols such as option identifiers are bound to their semantic content, and how these representations are ultimately converted into a final decision.

Prior MI work on MCQA has identified attention heads as key contributors to predicting the answer symbol: \citet{lieberum2023-mcqa} highlight ``Correct Letter Heads'' that attend from the final token to the correct symbol token, while \citet{wiegreffe2025-mcqa} localize symbol-predictive signals in mid-layer heads.
However, because these analyses intervene primarily at the last token position, they may miss binding dynamics if the model has already committed earlier.
Leveraging results on binding mechanisms in entity tracking \citep{feng2024-binding, dai2024-binding} tasks, \citet{mueller2025-mib} find evidence that models compute an index for the correct option and then dereference it to emit the corresponding symbol.

In this work, we focus on when and where models represent the correct option and its symbol.
We combine representational analyses (PCA and linear probing) with causal interventions (activation patching) to separate three stages of computation: per-option correctness evaluation, global option winner selection, and symbol binding at the answer-emission site.

Our contributions are as follows.
\begin{itemize}
    \item We perform representational analyses which show that models track both the position of symbols, and whether the current or previously seen option is correct. We further show that an option-boundary state (i.e., the newline after any MCQA option) often contains highly linearly decodable information related to the correctness of the option.
    \item Training 4-way winner-identity probes (predicting the symbol that the model outputs) and evaluating them under symbol permutations and content permutations yield a clear disassociation: 
    the winner \emph{content position} is already represented right after the model has processed the last option, while the identity of the \emph{symbol} emerges later.
    \item We find further support for this dissociation by performing causal interventions using probe-aligned patching. 
\end{itemize}

Together, these results suggest that language models use a two-stage mechanism in which they first commit to a winning option before they are prompted to respond and subsequently route/bind that winner tot he appropriate symbol closer to the point of answer emission.\label{sec:introduction}

\section{Related Work}
\paragraph{Symbol emission}
\citet{lieberum2023-mcqa} identify \emph{Correct Letter Heads}, attention heads that attend heavily from the final token to the correct label.
Low-rank analyses suggest that these heads may encode information about a letter (symbol) being the $n$-th item in a list.
As such, these heads promote the chosen symbol based on its position in the symbol ordering.
\citet{wiegreffe2025-mcqa} use vocabulary projection (\emph{logit lens}) and activation patching at the last token position to show that attention heads in middle layers play an important role in the prediction of an answer symbol.

\paragraph{Binding in language models}
Symbol binding has been a challenging problem in both neuroscience~\cite{burwick_binding_2014} and neural networks~\cite{von_der_malsburg_what_1999}.
A well-known example of binding can be found in \emph{entity tracking} \citep{kim2023-entity}, 
where a model must bind an entity to its attributes to later retrieve the correct attribute.
\citet{feng2024-binding} remark that these associations are made in context, so binding must occur in the activations of the language model instead of being stored as facts in the weights.
They propose the \emph{binding ID} (BI) mechanism: the model assigns abstract IDs to entities and attributes, and bound pairs share an ID that can be referenced later.
\citet{dai2024-binding} extend this line of work by capturing the \emph{ordering ID} (OI), the input order in which the entities and attributes appear in the context.
They show that OI is encoded in a low-rank activation subspace that influences in-context binding.

\noindent
Building on this, \citet{mueller2025-mib} argue that MCQA may rely on ordering IDs: the model identifies the answer's position in the option list and then dereferences that index to emit the corresponding symbol, echoing a hypothesis by \citet{lieberum2023-mcqa}.
\citet{lieberum2023-mcqa}'s preliminary experiments to analyze the outputs to the Correct Letter Heads suggest that models may use ``Content Gatherers'', which attend from the final tokens to the last token of the correct option content token.
In contrast, \citet{mueller2025-mib} report that order information shifts from the correct symbol token to the last token position in middle layers.
\label{sec:related_work}

\section{Experimental Setup}\label{sec:task_definition}
\paragraph{Task definition}
Following \citet{wiegreffe2025-mcqa} and \citet{mueller2025-mib}, we use a synthetic 4-way MCQA task to disentangle a model's task-specific knowledge from MCQA behavior.
Prior work relies on the Copying Colors from Context \citep[\textit{Colors};][]{wiegreffe2025-mcqa} dataset, adapted from the Memory Colors dataset introduced by \citet{norlund2021-colors}.
Its questions target prototypical colors of objects and are of the form ``A banana is yellow. What color is a banana?''
Because such facts may be stored in model weights,
the way questions are framed obscures the role of symbol binding, which is done
in context rather than retrieved \citep{feng2024-binding}.
To isolate MCQA behavior from factual recall, we modify the test set of the Colors dataset (4-way MCQA, 100 samples) by replacing the specific object names with ``\textit{this object}'', as in the following example.

\begin{promptbox}
{
\footnotesize
\texttt{Question: This object is white. What color is this object?}\\\\
\texttt{Options:}\\
\texttt{(A) white}\\
\texttt{(B) grey} \\
\texttt{(C) black} \\
\texttt{(D) brown}\\\\
\texttt{Answer: (}
}
\end{promptbox}

\noindent
Given that the color copying task is rather trivial--the correct option is explicitly mentioned in the prompt--we additionally experiment with the test set of the AI2 Reasoning Challenge Easy \citep[ARC-Easy;][]{clark2018-arc},
which consists of multiple multiple-choice elementary and middle-school science questions.
We restrict to four-option items (2365 samples in total), convert any numeric labels to letters, and use the same \texttt{Question}-\texttt{Options}-\texttt{Answer} prompt format as in the color-copying setup.

\paragraph{Models}
We experiment with two models: Llama-3.2-Instruct-3B \citep[28 layers;][]{meta2024-llama32, grattafiori2024-llama3} and Qwen3-8B \citep[36 layers;][]{yang2025-qwen3}.
\label{sec:experimental_setup}

\begin{figure}[t]
    \centering
    \includegraphics[width=\linewidth]{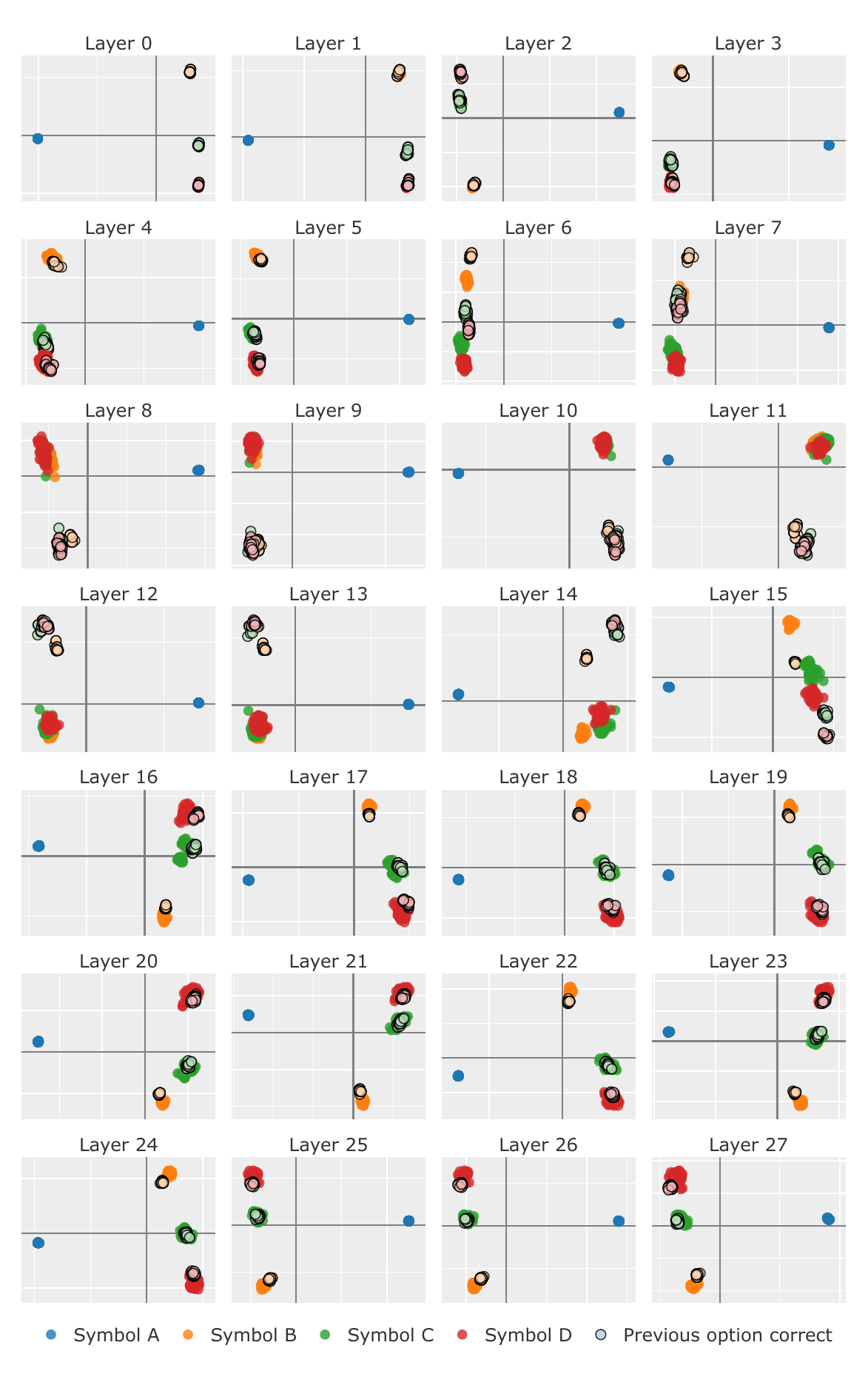}
    \caption{The first ($x$-axis) and second ($y$-axis) principal components from layerwise PCA in \textbf{Llama-3.2-3B}. Activations extracted at every \textbf{symbol} token position. The ``Previous option correct'' marker is in the color of the symbol of which the activations are being projected.}
    \label{fig:pca_symbols}
\end{figure}

\begin{figure}
    \centering
    \includegraphics[width=\linewidth]{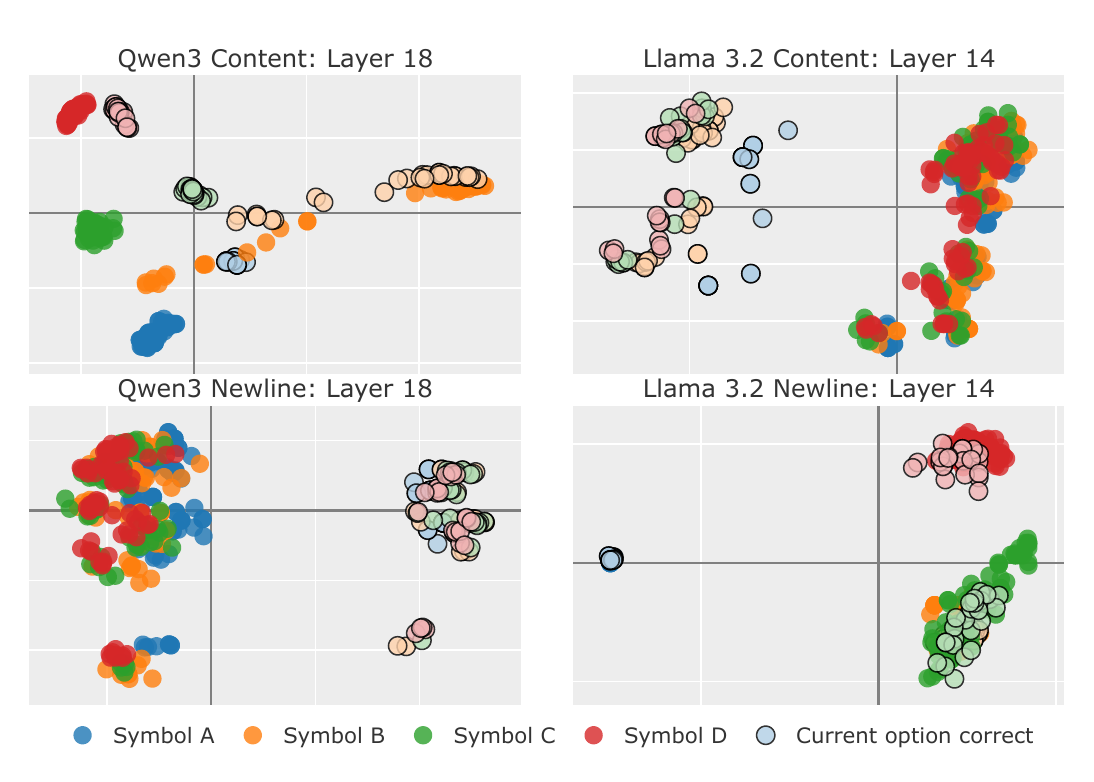}
    \caption{The first two principle components for Qwen3-8B (\textbf{left}) and Llama-3.2-3B-Instruct (\textbf{right}) when extracting the residual stream at the content tokens (\textbf{upper}) and subsequent newlines (\textbf{lower}). Note that, unlike Figure~\ref{fig:pca_symbols}, this plot tracks the {\em current}, not the {\em previous} token correct. Plots for all layers in Appendix~\ref{appendix:pca_content_newlines}.}
    \label{fig:pca_content_newline_subsets}
\end{figure}

\section{Finding Scoring Signals}

Following \citet{dai2024-binding}, we use principal component analysis (PCA) to find subspaces in the residual stream that encode the position of a symbol (such as {\em A, B, \textellipsis}) or content ({\em white, black, \textellipsis}).
For each MCQA prompt, we extract the $d$-dimensional activations at the token positions for each of the model's $n$ MCQA symbols 
and contents.
This results in an activation matrix $M \in \mathbb{R}^{n \times d}$,
for which the singular value decomposition can be written as $M = U \Sigma V^T$.
We focus on the first two principal directions in $V$.

\paragraph{Distinction between symbols}
Figure~\ref{fig:pca_symbols} shows the PCA for Llama-3.2-3B-Instruct (see Appendix~\ref{app:qwen-pcs} for Qwen-8B). Activations cluster by symbol position, indicating that models internally distinguish positional information.
In particular, the first principal component separates the first option from later ones.
To test whether this effect is merely positional, or tied to the literal symbol “A”,
we permute option labels and re-run PCA (Appendix~\ref{appendix:pca}, Figures \ref{fig:appendix_pca_llama_permuted_symbols} and \ref{fig:appendix_pca_qwen_permuted_symbols}).
For Qwen3-8B, this permutation substantially alters the structure of the representation space, suggesting an entanglement between positional information with the semantic (alphabetical) ordering of the option symbols.
In contrast, the representations of Llama-3.2-3B-Instruct appear to be more strongly driven by the ordering of options in the list.

In both models, the first option remains atypical.
One explanation is that, when processing the first element of a list, the lists consists of a single element and no comparison with other items is possible.
Consequently, the representation may not yet encode an explicit ``$n$-th item'' signal.
A similar effect appears in the OI subspace visualizations of \citet{dai2024-binding}, where the first item behaves differently despite the existence of an overall low-rank ordering direction.

\paragraph{Tracking of correctness}
The first option may look different because the model could track whether a previously seen option is correct.
In an autoregressive language model, options are processed sequentially, so any ``winner-so-far'' signal must be stored \emph{after} the option deemed correct; the first position cannot encode a prior-correctness trace.
The idea that a symbol may store information about a \emph{previous} answer has also been briefly suggested by \citet{lieberum2023-mcqa}.

\noindent
Figure~\ref{fig:pca_symbols} supports this ``correctness tracking'' hypothesis:
within each symbol cluster, there is further separation of points
depending on whether the preceding option was correct (e.g., "B" correct while processing "C").
The effect is most pronounced in middle layers, which have been associated with feature construction \citep{lad2024-inference}.
Figure~\ref{fig:pca_content_newline_subsets} suggests that this ``correctness'' signal may arise even earlier, at the option content token or the newline marking the option boundary.

\paragraph{ARC-Easy}
Given the trivial nature of the color copying task, the apparent separation between the correct and incorrect answers in Figure \ref{fig:pca_symbols} and \ref{fig:pca_content_newline_subsets} may simply reflect whether a given color appeared in the question, rather than whether the option is actually correct.
In other words, the signal may be tracking an option's presence in the prompt instead of its correctness.
We repeat the subspace visualization using the ARC-Easy dataset.
Given that ARC-Easy is less structured than the color copying dataset, it is to be expected that its patterns are less readily available in PCA space. That said, we can identify layers in Figures \ref{fig:appendix_pca_llama_arc} and \ref{fig:appendix_pca_qwen3_arc} (Appendix~\ref{appendix:pca}) that suggest some degree of separability between correct and incorrect options.\label{sec:experiments_pca}

\section{Probing Correctness Scores}
PCA offers a qualitative picture of how correctness-related variation may be structured in the residual stream: across layers and token positions, activations exhibit low-dimensional patterns consistent with a correct/incorrect separation.
Based on this exploratory analysis, and noting that PCA captures
directions of maximal variance that need not align with task-relevant variables, we perform more targeted tests.
We use linear probes \citep[][aka \emph{diagnostic classifiers}; \citet{hupkes2018-diagnostic}]{alain2017-probing} to quantify correctness/scoring effects suggested by our PCA visualizations.

\paragraph{Notation}
Let $l \in \{0, \dots, L\}$ be the layers of a language model and $d$ the hidden size.
For each prompt $i$ and option $X \in \{A, B, C, D\}$, let $p_{i, X}$ be a token position associated with that option (e.g., the symbol token, or an option-content token).
Let $r_{i, X}^{(l)} \in \mathbb{R}^d$ be the residual stream vector at layer $l$ at  position $p_{i, X}$.
For each option $X$ in sample $i$, we define $y_{i, X} = 1$ if $X$ is the correct option in sample $i$ and  $y_{i, X} = 0$ otherwise.
A linear probe at layer $l$ is a parameter vector $w^{(l)} \in \mathbb{R}^d$ and bias $b^{(l)} \in \mathbb{R}$, which defines a classifier 
$\hat{y}_{i, X} = \sigma((w^{(l)})^T r^{(l)}_{i, X} + b^{(l)})$, where $\sigma(\cdot)$ is the logistic sigmoid.
Hence, a learned linear function $w^T r + b$ that best predicts whether an MCQA option is correct according to the language model.

We train linear probes (on 90\% of ARC-Easy; 10\% for testing) to address
(1) whether the model encodes \emph{per-option correctness} as it processes the options; and (2) whether it represents the winning content and eventual output symbol \emph{before} the final answer-emission token.

\begin{figure}
    \centering
    \includegraphics[width=\linewidth]{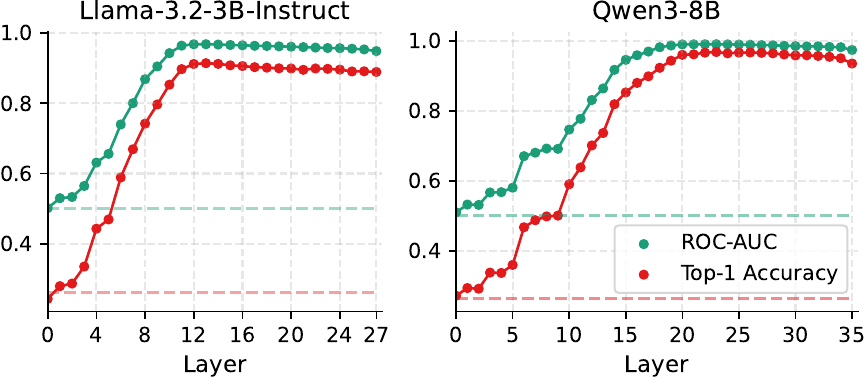}
    \caption{Layer-wise performance of \textbf{binary per-option probes} trained for the \textbf{ARC-Easy} dataset with $C=10^{-4}$ on the residual stream. The dashed lines represent random chance baselines. 
    }
    \label{fig:probing_binary}
\end{figure}
 
\subsection{Per-option Correctness}
To test whether \emph{per-option correctness} is linearly decodable from the model's internal representations,
for each MCQA prompt we extract the residual-stream activation at the end-of-option boundary (e.g., \texttt{\textbackslash n}, \texttt{{.\textbackslash n}}, or \texttt{\textbackslash n\textbackslash n}) and train an L2-regularized logistic-regression probe to predict whether the option is the one the model will select as correct.
Probes are trained on ARC-Easy to reduce the risk that apparent signals reflect trivial lexical overlap between the question and an option. We evaluate on held-out ARC-Easy and the synthetic color-copying set.
Because options from the same question are correlated, we use question-ID grouped splits and tune the inverse regularization strength $C$ via 10-fold GroupKFold cross-validation.

\paragraph{Results}
To account for the 1:3 class imbalance (one correct vs.\ three incorrect options per question), we use ROC-AUC as our primary metric.
For each held-out question, we score all four options using the probe and also predict the answer as the option with the largest probe score;
top-1 accuracy is then the fraction of questions for which this operation recovers the true correct option (the language model's output).
This metric captures whether the probe's scores are informative \emph{within} each question, rather than only in aggregate across the dataset.

\noindent
Figure~\ref{fig:probing_binary} shows that both AUC and top-1 accuracy improve with depth: probes are weak in early layers but rise through the middle and remain strong late in the network.
Compared to PCA (Section~\ref{sec:experiments_pca}), which reveals a correctness separation
correctness in intermediate layers, probing indicates that correctness remains linearly decodable even in final layers.
A plausible interpretation is that late layers reorganize the representation:
rather than maintaining correctness as a dominant global axis of variation, the model may transform it into a more ``decision-like'' representation.
Under such a reorganization,
unsupervised PCA projections can understate separability even as the supervised decoding task becomes easier.

\noindent
The ARC-Easy probe direction also transfers to the color copying task.
Freezing $w$ and scoring newline activations with $s(x) = w^T x$ yields near-perfect separation:
option-level AUC and top-1 accuracy both reach 1.0 across middle-to-late layers (layers 12-27 for Llama-3.2-3B-Instruct;layers 20-35 for Qwen3-8B).
Overall, this indicates a robust per-option ranking signal, a structured ``option evaluation'' feature in the residual stream that generalizes across MCQA datasets with the same format.

\begin{figure}
    \centering
    \includegraphics[width=\linewidth]{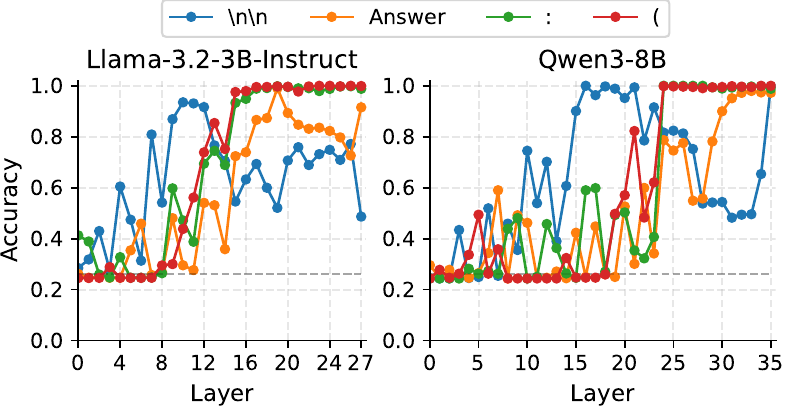}
    \caption{Layer-wise performance of \textbf{multinomial probes trained} for the \textbf{ARC-Easy} dataset with $C = 10^{-2}$. Each line corresponds to token positions of  \texttt{\textbackslash n\textbackslash nAnswer:~(}, right after the model has processed all the MCQA options. Dark gray dashed lines represent random chance baselines.}
    \label{fig:probing_multi}
\end{figure}

\subsection{Winner Identity Probe}
To test whether the model consolidates option-level evaluations into a single global decision prior to answer emission, we train a 4-way winner identity probe.
For each prompt, we label the correct option index $y \in \{A, B, C, D\}$ and train a multinomial linear classifier on residual stream activations that occur after \emph{all} options have been processed: the newlines after the final option (the \emph{end-of-options checkpoint}) and the \texttt{Answer: (} prefix.
By sweeping layers and checkpoints, we estimate when winner identity becomes linearly decodable: decodability at the end-of-options checkpoint supports \emph{early commitment}, whereas emergence only near answer emission supports late consolidation or binding.

\paragraph{Results}
For both Llama-3.2-3B-Instruct and Qwen3-8B, winner identity exhibits a consistent two-stage progression (Figure~\ref{fig:probing_multi}).
First, it becomes reliably decodable at the end-of-options checkpoint in intermediate layers, reaching high accuracy before the model encounters the answer prefix.
Second, decodability at answer prefix tokens remains weak until later layers, where it rises sharply and often saturates.
This pattern is consistent with winner information being available early in the computation but only being made accessible at specific positions that directly govern the next-token distribution over answer symbols.

\noindent
The contrast across checkpoint tokens refines this interpretation.
The end-of-options boundary tends to show earlier decodability, while the \texttt{Answer:\ (} region shows the strongest late-layer decodability, which is in line with its role as immediate context for symbol emission.
The end-of-option newline curves being non-monotonic suggests re-localization rather than simple accumulation: once winner identity has been actively written into the answer region by attention heads, it would no longer need to remain cleanly separable at this earlier checkpoint.
Figure~\ref{fig:appendix_multiprobe_colors} (Appendix~\ref{appendix:probe}) shows that these results, too, are generalizable to the color copying task.

\begin{figure*}[t]
\centering
\begin{subfigure}[b]{0.49\textwidth}
    \centering
    \includegraphics[width=\linewidth]{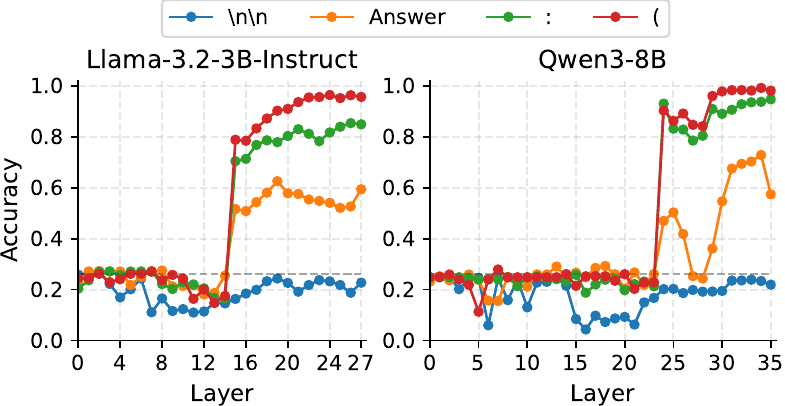}
    \caption{Probing the correct \textbf{symbol}}\label{fig:multiprobe_permuted_symbol}
\end{subfigure}
~
\begin{subfigure}[b]{0.49\textwidth}
    \centering
    \includegraphics[width=\linewidth]{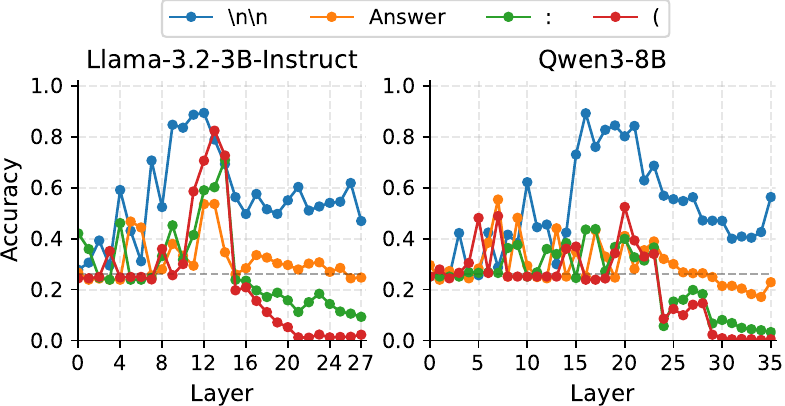}
    \caption{Probing the correct \textbf{content}}\label{fig:multiprobe_permuted_content}
\end{subfigure}

\caption{Performance of the multinomial probe from Figure~\ref{fig:probing_multi} on ARC-Easy prompts in which option \textbf{symbols have been permuted}. We probe for \textbf{(a)} the correct output symbol under this permutation as well as for \textbf{(b)} the correct (unchanged) content position.}\label{fig:multiprobe_permuted}
\end{figure*}

\noindent
We further test whether the probes are reading out a symbol-coded decision (the symbol/letter to emit) or a content-coded decision (which option content or position is correct) using symbol-permutation generalization without retraining the probes (Figure~\ref{fig:multiprobe_permuted}).
Concretely, for each prompt we randomly reassign the option symbol while leaving the order of the textual options unchanged;
this breaks the alignment between ``the winning option content'' and ``the winning symbol.''
We then evaluate the same permuted prompts under two target definitions.
When the targets reflect the new correct (permuted) symbol, (Figure~\ref{fig:multiprobe_permuted_symbol}), performance at the end-of-options checkpoint collapses across all layers, while accuracy in the answer region increases late and approaches ceiling.
Conversely, when the targets are (we probe for) the content (position) of the winner (Figure~\ref{fig:multiprobe_permuted_content}), performance remains strong at end-of-options checkpoints in middle layers but collapses near emission.
Together, these results support a two-stage mechanism: the model first selects a winner in option/content space immediately after reading the candidates, then performs a late binding/routing step during the answer prefix that maps the winner identity onto the correct output symbol under the current symbol-content assignment.
\label{sec:experiments_probing}

\section{Causal Interventions}
To understand whether the signals found in previous sections are \emph{causally relevant}, we perform a series of activation patching experiments.
\emph{Activation patching} \citep{vig2020-patching, meng2022-patching, geiger2020-patching, wang2023-patching} is a widely used in MI to causally attribute specific model behavior to model components (e.g., specific attention heads).
It requires two prompts: a \emph{clean} prompt, on which the model exhibits some desired behavior, and a \emph{corrupted} (or \emph{counterfactual}) prompt that is a modified version of the original prompt.
The model is first run on both prompts, and the activations (e.g., MLP output) of the clean prompt are cached.
The model is then re-run on the corrupted prompt while a chosen subset of its activations are patched with those from the clean run.\footnote{Note that it is also possible to patch activations from a corrupted prompt into a clean run without it necessarily being the symmetric case of patching from the clean into the corrupted prompt \citep{zhang2024-patching}.}
By comparing the output of the patched corrupted run with the original corrupted run, we can quantify the causal contribution of the selected components to preserving or restoring the clean behavior.
It must be noted that in the case of activation patching, the corrupted/counterfactual prompt essentially defines the task \citep{miller2024-metrics}.
In our experiments, we make use of two kinds of counterfactuals, either \em{swapping the correct symbol} with a different one, or \em{swapping the correct content}.

\paragraph{Metric}
We evaluate our patching results using the \emph{faithfulness} metric \citep{wang2023-patching} averaged over all samples:
\begin{equation}
\footnotesize
    \text{Faithfulness} = \frac{\ell_\text{patched} - \ell_\text{corrupt}}{\ell_\text{clean} - \ell_\text{corrupt}}\text{,}
\end{equation}
where $\ell_x$ denotes the logit for the clean answer token in run $x$.

\subsection{Binary Probe Patching}
To connect the representational probes from Section~\ref{sec:experiments_probing} to causal tests, we perform \emph{probe-aligned patching}, a targeted activation-editing procedure that directly intervenes on a one-dimensional subspace of the residual stream.
Let $w$ be a unit-norm probe direction, and consider a clean and corrupted run.
At layer $l$, position $i$, and option $X$, define:

\begin{align*}
    s_\text{clean} = w^Tr^{(l)}_{i, X, \text{clean}}\text{,} \qquad s_\text{corr} = w^Tr^{(l)}_{i, X, \text{corrupt}}\text{.}
\end{align*}

\noindent Probe-aligned patching replaces only the probe-aligned component of the corrupted residual stream with the corresponding value from the clean run, while leaving the orthogonal complement unchanged.
For the binary probes, this means:
\begin{equation}
\label{eqn:binary}
    r^{(l)}_{i, X, patched} = r^{(l)}_{i, X, \text{corrupt}} + (s_\text{clean} - s_\text{corr}) w\text{.}
\end{equation}
This update is the unique minimal-norm modification that enforces $w^Tr^{(l)}_{i, X, patched} = s_\text{clean}$.
It matches the clean run's coordinate along $w$ while preserving all other components of $r^{(l)}_{i, X, \text{corrupt}}$.
Intuitively, this intervention ``edits'' the amount of the probed feature present in the corrupted activation without overwriting unrelated information.

\paragraph{Patch positions}
When we patch activations at the newline immediately following an option's content, there are therefore two natural ways to decide where a clean option's activation should be inserted into the counterfactual prompt: \emph{content-aligned} patching, which patches a clean option into the counterfactual option that contains the same content, and \emph{symbol-aligned} patching, which patches a clean option into the counterfactual option that carries the same symbol.
We apply the chosen alignment consistently to both the true (clean) winner and the option it is swapped with, preserving their relative relationship under the intervention.
Intuitively, content-aligned patching tests whether the relevant signal is attached to the option's content as it moves across labels, while symbol-aligned patching tests whether the signal is attached to the symbol itself.

\begin{table*}
    \centering
    \setlength{\tabcolsep}{5pt}
    \small
    \begin{tabular}{lcccc}
        \toprule
        &\multicolumn{2}{c}{\textbf{Llama-3.2-3B-Instruct}} & \multicolumn{2}{c}{\textbf{Qwen3-8B}} \\
        \cmidrule(lr){2-3} \cmidrule(lr){4-5}
        & \textbf{Moved symbol} & \textbf{Moved content} & \textbf{Moved symbol} & \textbf{Moved content}  \\\midrule
        \textbf{Content-aligned} & $-0.01$ $(1.02)$ & $-0.281$ $(0.702)$ & $0.047$ $(0.760)$ & $-0.003$ $(0.713)$ \\
        \textbf{Symbol-aligned} & $0.870$ ($0.937$) & $0.747$ ($0.178$) & $0.450$ $(0.717)$ & $0.380$ ($0.910$)\\\bottomrule
    \end{tabular}
    \caption{Faithfulness scores of option-boundary patching under symbol- vs.\ content-aligned probe-aligned patching. Numbers between parentheses indicate scores for full residual vector patching.} \label{tab:llama_binary_patch}
\end{table*}

\paragraph{Results}
Table~\ref{tab:llama_binary_patch} compares the content-aligned and symbol-aligned patching results.
Across both counterfactual regimes, probe-aligned patching of the 1D ``correctness'' feature is markedly more faithful under symbol-aligned correspondence than under content alignment, indicating that this low-dimensional feature transfers most effectively when matched by the symbol rather than by the option content.
Full-vector patching at the same newline positions reveals a complementary structure: when the content is permuted, content-aligned full patching is substantially higher than symbol-aligned, consistent with the full residual stream at option boundaries containing rich information about the content.
Notably, however, in the symbol-aligned moved-content case for Llama-3.2-3B-Instruct, full-vector patching is much less faithful than probe-directed patching.
This suggests that overwriting the entire residual vector at a symbol-matched site introduces substantial mismatch with the surrounding context, whereas selectively adjusting a single probe-aligned coordinate can preserve most local structure while still reinstating the particular feature that drives behavior on the clean prompt.
Intuitively, these patterns point to option-boundary representations that mix high-dimensional content-specific state with a lower-dimensional, symbol-coupled component.

\noindent
This could reflect a local routing feature that later attention uses to bind the selected winner to the emitted symbol.
Across the two models, the same intervention reveals different indexing structures at option boundaries: in Llama-3.2-3B-Instruct, option-boundary states appear content-context-sensitive (full vector replacement into content-mismatched positions is harmful), whereas in Qwen3-8B they appear more amenable to symbol-aligned transplantation.
This suggests model-dependent variation in how option summaries are stores and later routed into the answer-emission region, without requiring that the output symbol be explicitly represented at the end-of-option token.

\subsection{Multinomial Probe Patching}
Let $W \in \mathbb{R}^{4 \times d}$ be the weight matrix of our mulitnomial probe, and $P \in \mathbb{R}^{d \times d}$ the orthogonal projector onto the row space of $W$, i.e., $P = UU^T$ for an orthonormal basis $U$ spanning $\text{Row}(W)$. Then patching in the residual stream (analogous to Eq~(\ref{eqn:binary}) for the binary probe) is defined as 
\begin{equation}
    \footnotesize
    r_{i, X, patched}^{(l)} = r_{i, X, \text{corr}}^{(l)} + P(r_{i, X, \text{clean}}^{(l)} -  r_{i, X, \text{corr}}^{(l)})\text{.}
\end{equation}

\begin{figure}
    \centering
    \includegraphics[width=0.8\linewidth]{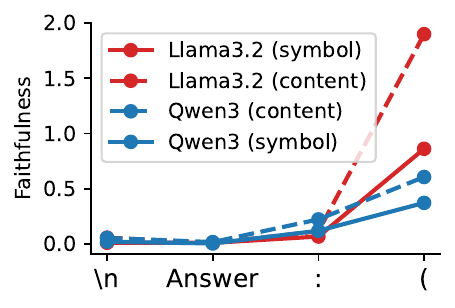}
    \caption{Faithfulness scores of multinomial winner identity probe patching under symbol and content perturbations. Dashed lines display results from using content-perturbed counterfactuals whereas solid lines are generated under symbol perturbations.}
    \label{fig:4way_faithfulness}
\end{figure}

\paragraph{Results}
In Figure~\ref{fig:4way_faithfulness}, faithfulness is negligible when patching at the end-of options checkpoint or the ``\texttt{Answer}'' prefix, but increases sharply at the final pre-emission token \texttt{(}.
Thus, although correctness-related structure is detectable earlier in the residual stream, the probe-aligned representation that directly controls the answer token logits becomes causally effective only near emission.
When faithfulness exceeds 1
(at the \texttt{(} token) reflects overshoot, 
patched runs assign the clean answer token a higher logit than the clean run, consistent with probe-subspace interventions isolating a strongly behaviorally-aligned component.

\noindent
This pattern supports a two-stage account: First,
during option processing, models compute and store local evaluation variables at boundary positions between options.
Signals at the end-of-options boundary are linearly decodable, but patching evidence in Figure~\ref{fig:4way_faithfulness} indicates that they are not, by themselves, the control variables that determine the emitted answer token.
Second,
after the model encounters the answer prefix, it may convert its option-level information into a global, answer-controlling representation that is directly coupled to next-token prediction.
The increase at the colon \texttt{:} and \texttt{(} is consistent with this conversion: the answer prefix cues a routing step that consolidates the decision into a representation whose geometry aligns with the multinomial probe's row space and whose value is immediately used by the unembedding to determine the next-token distribution.\label{sec:experiments_patching}

\section{Discussion}
By combining representational measurements and visualizations with causal interventions, we find consistent evidence for a two-stage computation in language models when they perform multiple-choice question answering.
First, models represent option-level evaluations and consolidate a winner in content space shortly after having processed the final option.
Second, closer to answer emission, models route that winner to the appropriate output symbol.

Our results are broadly compatible with prior MI studies on MCQA, especially the finding that answer-symbol information becomes highly accessible near emission.
However, our account refines the interpretation in two ways.
First, by probing and intervening at earlier positions in the prompt sequence, and not just the final token position, we show that important decision-relevant computation often occurs before the model reaches the emission site.
Winner identity in content space becomes decodable immediately after the final option is processed, supporting an \emph{early commitment} regime that analyses focusing solely on the last token may miss.
Second, by using symbol permutations as controlled perturbations, we disambiguate whether this winner identity is represented as a symbol-coded decision or a content-coded decision.
This yields a clear dissociation: intermediate representations track the winner \emph{content} position \cite[in line with findings on content ordering IDs by][]{mueller2025-mib}, while the explicit symbol identity emerges later, near answer time. These findings tease apart processes that are conflated in previous work, namely (a) 
the model selecting the correct answer; and (b) the model selecting the symbol to emit. An interesting question for future work is whether these two processes can also be distinguished using mechanistic interventions on attention heads, in addition to the residual stream.\label{sec:discussion}

\section*{Limitations}
This work aims to characterize how instruction-tuned language models solve MCQA prompts, with an emphasis on separating option evaluation, winner selection, and symbol binding.
While the results are generally consistent across the two models we study, several limitations are worth noting.

\paragraph{Scope of models}
We focus on relatively small models from two specific model families.
Mechanistic findings, especially localization claims about when and where particular signals are represented, can vary with architecture, model scale and post-training procedures (e.g., supervised fine-tuning).

\paragraph{Prompt and task dependence}
Our experiments use a specific MCQA formatting template for two datasets (ARC-Easy and a synthetic color copying task) designed to control lexical confounds and cleanly separate symbol/content manipulations.
Different datasets, option formatting conventions and response constraints could shift representational geometry and causal pathways.
In particular, models may adopt different strategies when prompted for explanations, chain-of-thought, or direct answer text rather than a single symbol.

\paragraph{Known MCQA biases}
We do not evaluate documented MCQA artifacts (length biases, position bias, etc.).
Instead, our goal is mechanistic decomposition and consequently, additional work is required to quantify how much each artifact is explained by binding versus other factors such as lack of knowledge.

\paragraph{Interpretation of probes and their causal interventions}
Linear probes provide evidence that particular variables are linearly decodable from activations, but they do not by themselves establish that the model \emph{uses} those variables in that form.
In our work, too, the standard interpretability caution remains: high probe performance can reflect correlates or downstream consequences rather than functional intermediate representations.
Additionally, probe performance depends on dataset balance, split strategy, and regularization; while we used grouped splits and cross-validation to select the inverse regularization term, subtle distributional differences in the dataset can influence absolute performance.

In the binary per-option correctness patching experiments, we intervene along a single probe direction (a 1D manipulation), which may be too restrictive if the causally relevant ``option evaluation'' representation is distributed across multiple dimensions.
In contrast, in our winner-identity analyses we effectively consider a multi-class linear readout, which corresponds to a higher-dimensional subspace that can capture richer structure.
This asymmetry means that differences in intervention success should not be over-interpreted as differences in ``causal reality'' of the underlying signals without accounting for the capacity of the intervention itself.
More systematic comparisons (e.g., varying the dimensionality of probe-aligned interventions, or matching intervention rank across binary and multinomial settings) would better isolate whether failures arise from insufficient intervention dimensionality versus the absence of a causal signal.

\bibliography{refs}

\clearpage
\appendix
\section{Probing}
Hyperparameter search for the inverse regularization strength was done over a grid $[0.0001, 0.001, 0.01, 0.1, 0.3, 1, 3]$, with a maximum of 5000 iterations. We used the \texttt{liblinear} solver for the binary probes and \texttt{lbfgs} for the multinomial probes. All probes were trained using \texttt{scikit-learn}.

\begin{figure}
    \centering
    \includegraphics[width=\linewidth]{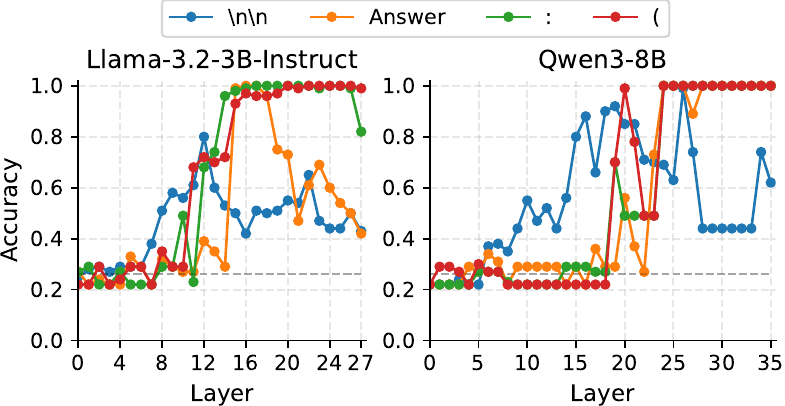}
    \caption{Performance on the color copying task of the multinomial probes trained on ARC-Easy (without additional re-training on the color copying dataset).}
    \label{fig:appendix_multiprobe_colors}
\end{figure}

Figure~\ref{fig:appendix_multiprobe_colors} shows performance of the probe on the color copying task after training on ARC-Easy.
\label{appendix:probe}

\label{appendix:tokenization}
\section{PCA Results}
In this section, we present additional results from our PCA experiments.

\subsection{PCA Qwen3-8B: Symbol Positions}\label{app:qwen-pcs}
\begin{figure}
    \centering
    \includegraphics[width=\linewidth]{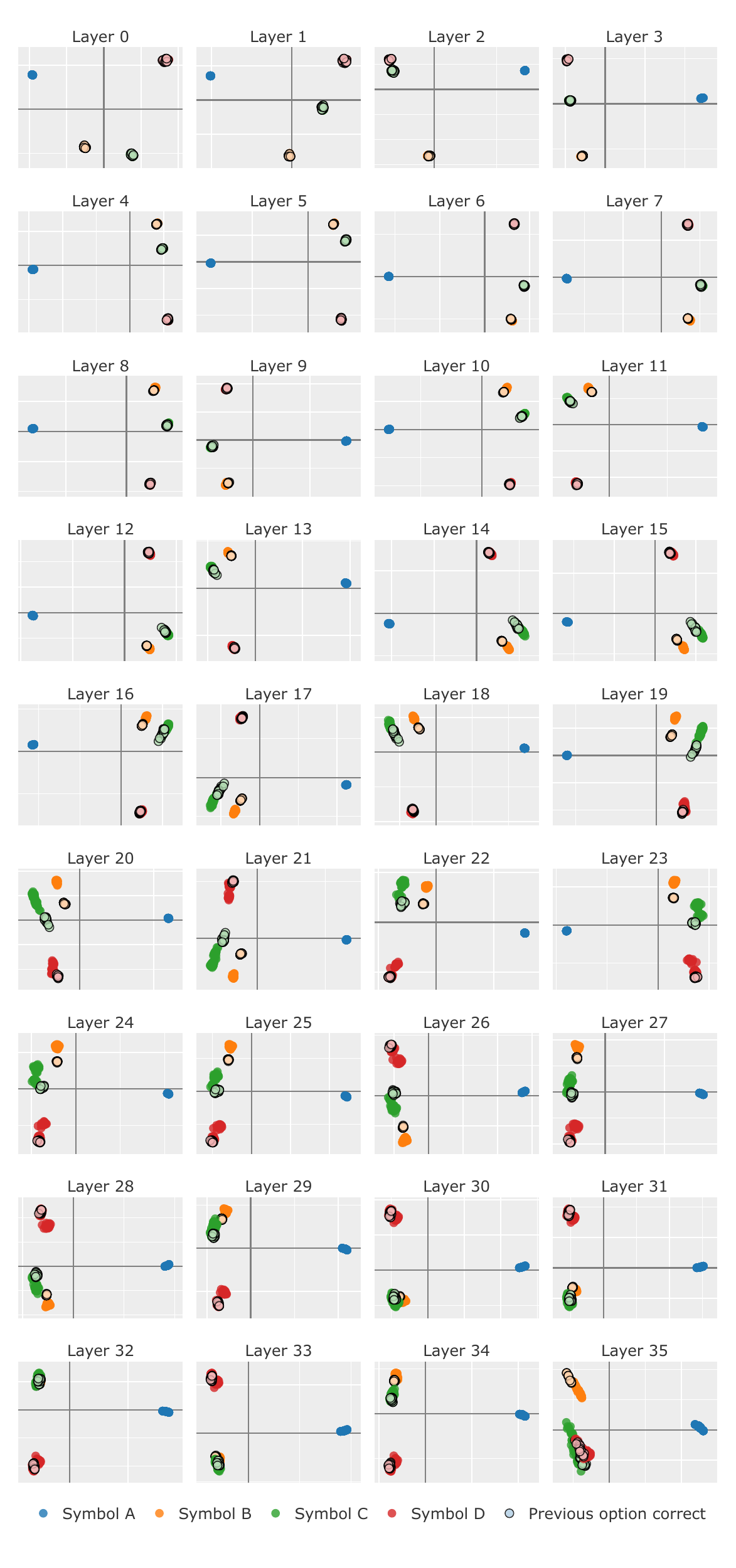}
    \caption{PCA for activations extracted at the \textbf{symbol positions} for \textbf{Qwen3-8B}, analogous to Figure~\ref{fig:pca_symbols} shown for Llama-3.2-3B in Section~\ref{sec:experiments_pca}.}
    \label{fig:appendix_pca_qwen3_symbols}
\end{figure}

Figure~\ref{fig:appendix_pca_qwen3_symbols} shows PCA for activations at symbol positions for Qwen3-8B.

\subsection{PCA with Permuted Symbols}
\begin{figure}
    \centering
    \includegraphics[width=\linewidth]{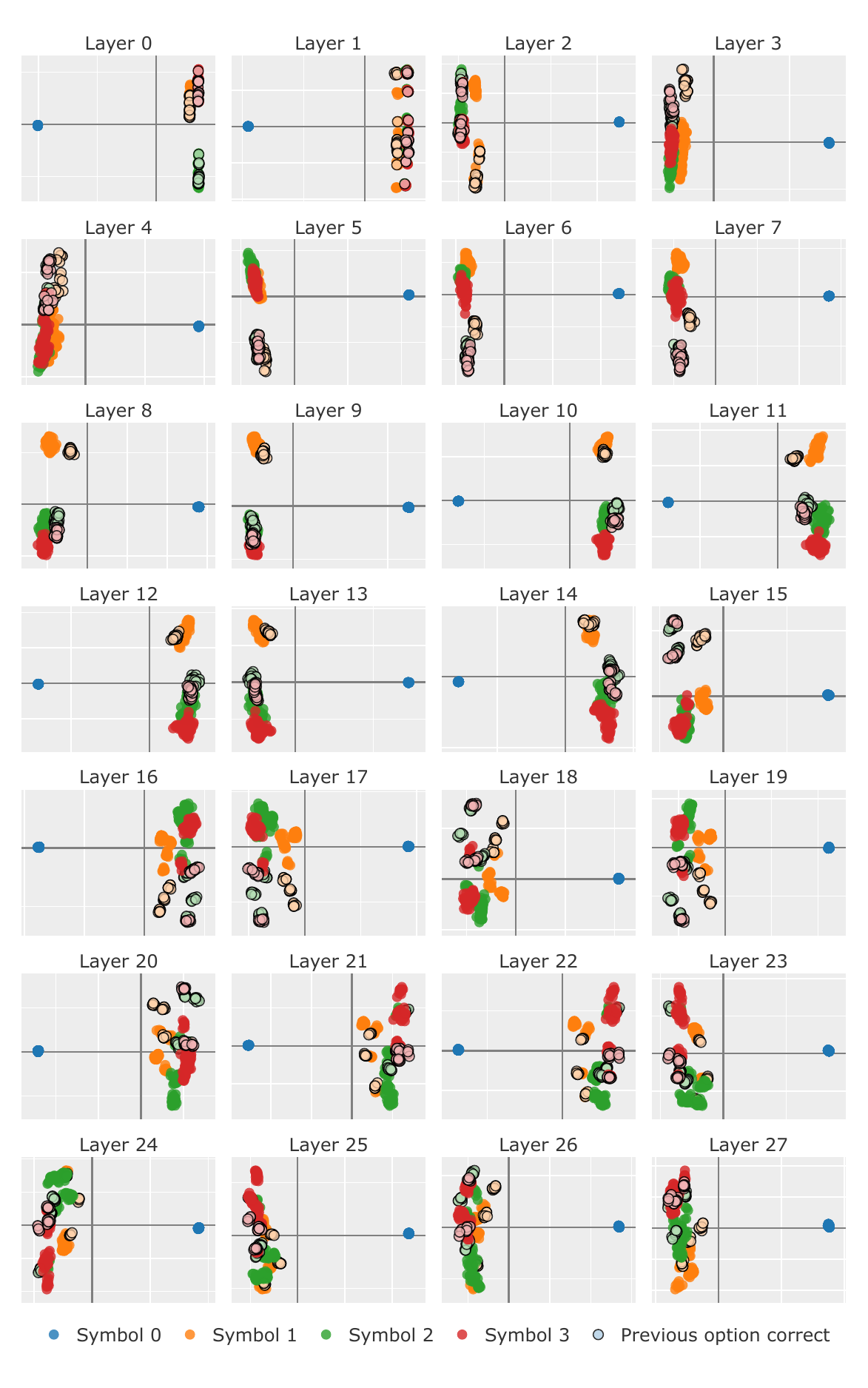}
    \caption{PCA for activations of \textbf{Llama-3.2-3B-Instruct} extracted at the \textbf{symbol positions} when the symbols in the prompt are \textbf{permuted}.}
    \label{fig:appendix_pca_llama_permuted_symbols}
\end{figure}

\begin{figure}
    \centering
    \includegraphics[width=\linewidth]{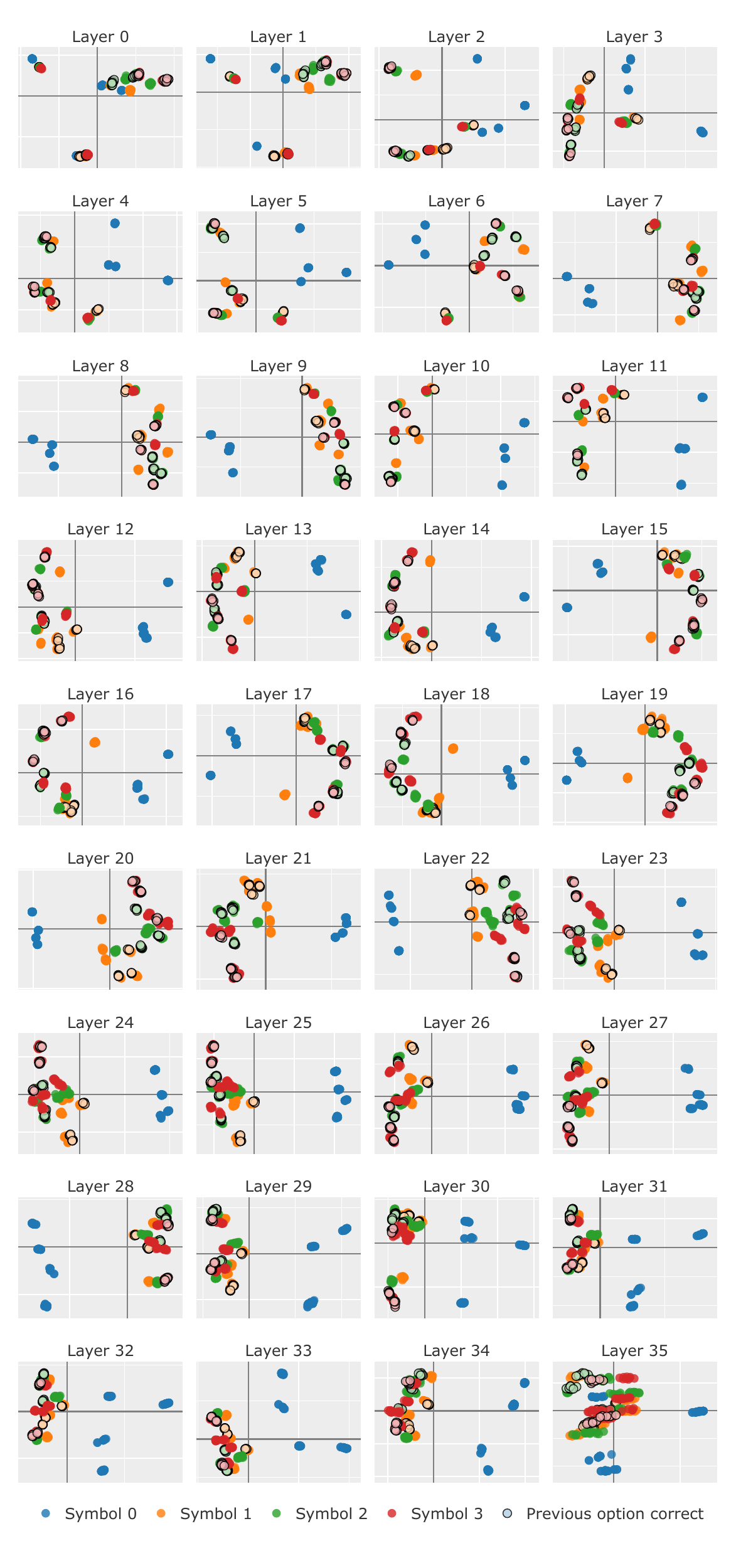}
    \caption{PCA for activations of \textbf{Qwen3-8b} extracted at the \textbf{symbol positions} when the symbols in the prompt are \textbf{permuted}}
    \label{fig:appendix_pca_qwen_permuted_symbols}
\end{figure}

Figures~\ref{fig:appendix_pca_llama_permuted_symbols} and \ref{fig:appendix_pca_qwen_permuted_symbols} show PCA for activations at symbol positions after prompt symbols are permuted.

\subsection{PCA: Content and Newline Positions}\label{appendix:pca_content_newlines}
\begin{figure}
    \centering
    \includegraphics[width=\linewidth]{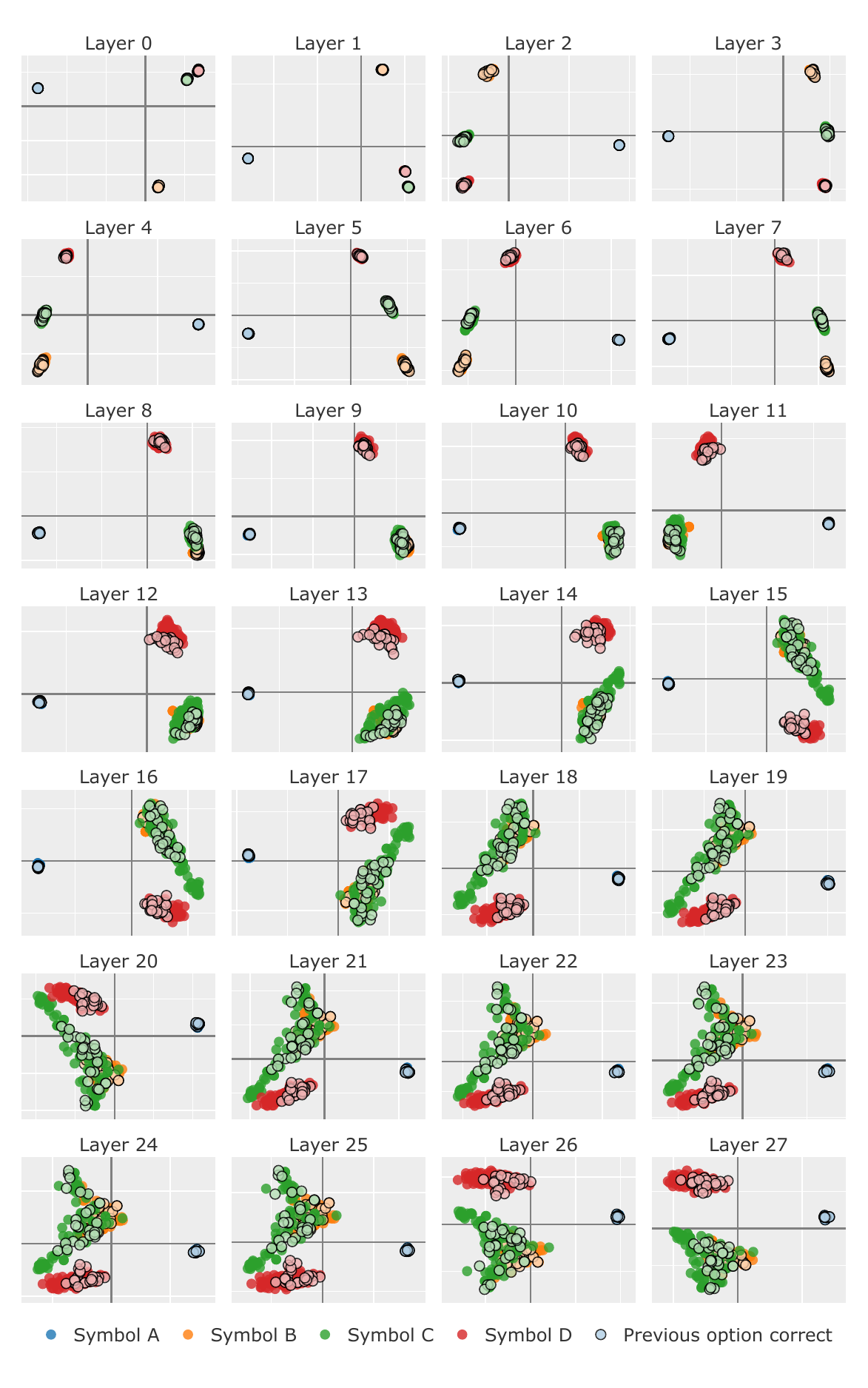}
    \caption{PCA for hidden activations extracted from \textbf{Llama-3.2-3B-Instruct} at the \textbf{content token} positions.}
    \label{fig:appendix_pca_llama_content}
\end{figure}

\begin{figure}
    \centering
    \includegraphics[width=\linewidth]{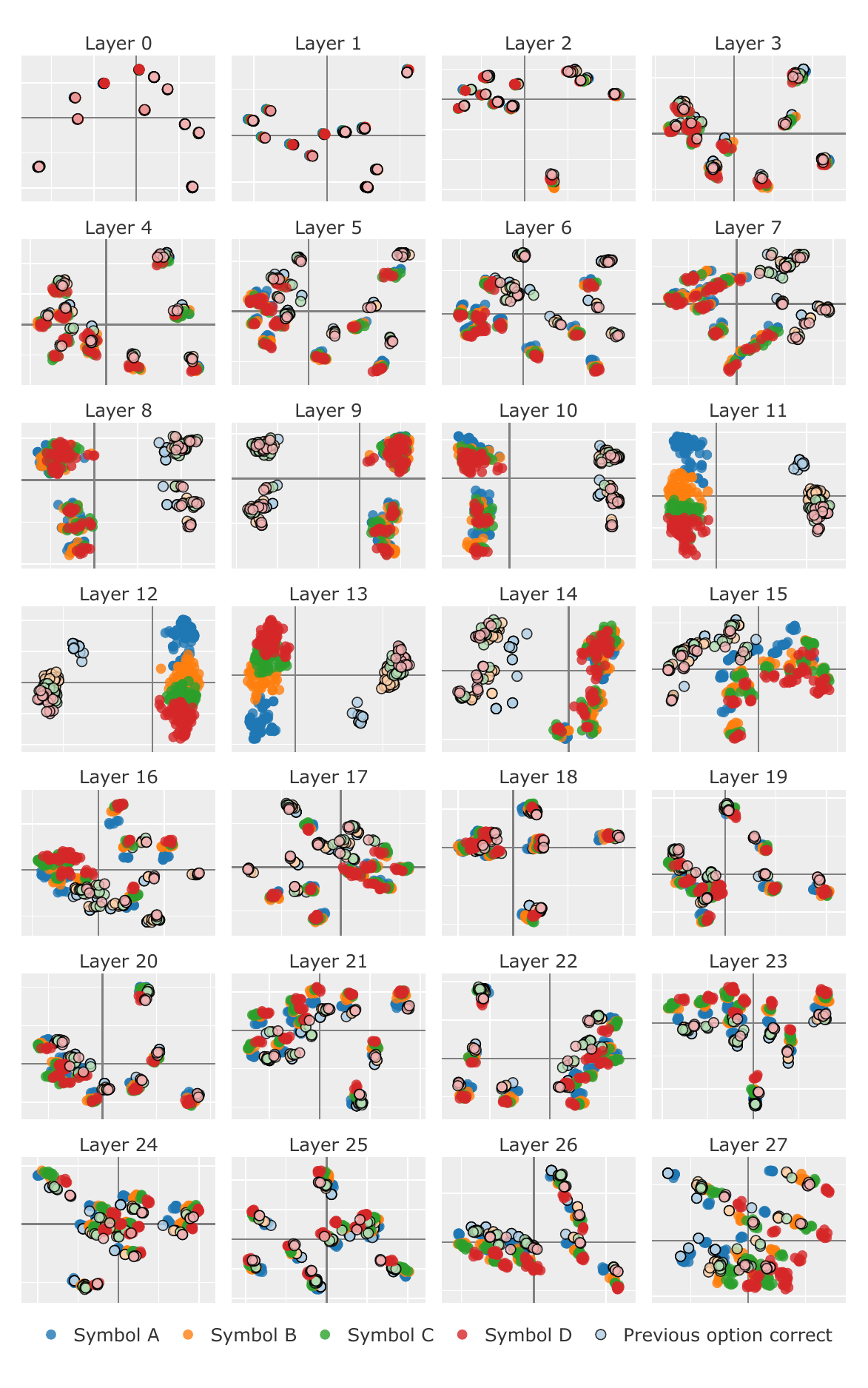}
    \caption{PCA for hidden activations extracted from \textbf{Llama-3.2-3B-Instruct} at the \textbf{newline tokens} directly after the content tokens.}
    \label{fig:appendix_pca_llama_newline}
\end{figure}

\begin{figure}
    \centering
    \includegraphics[width=\linewidth]{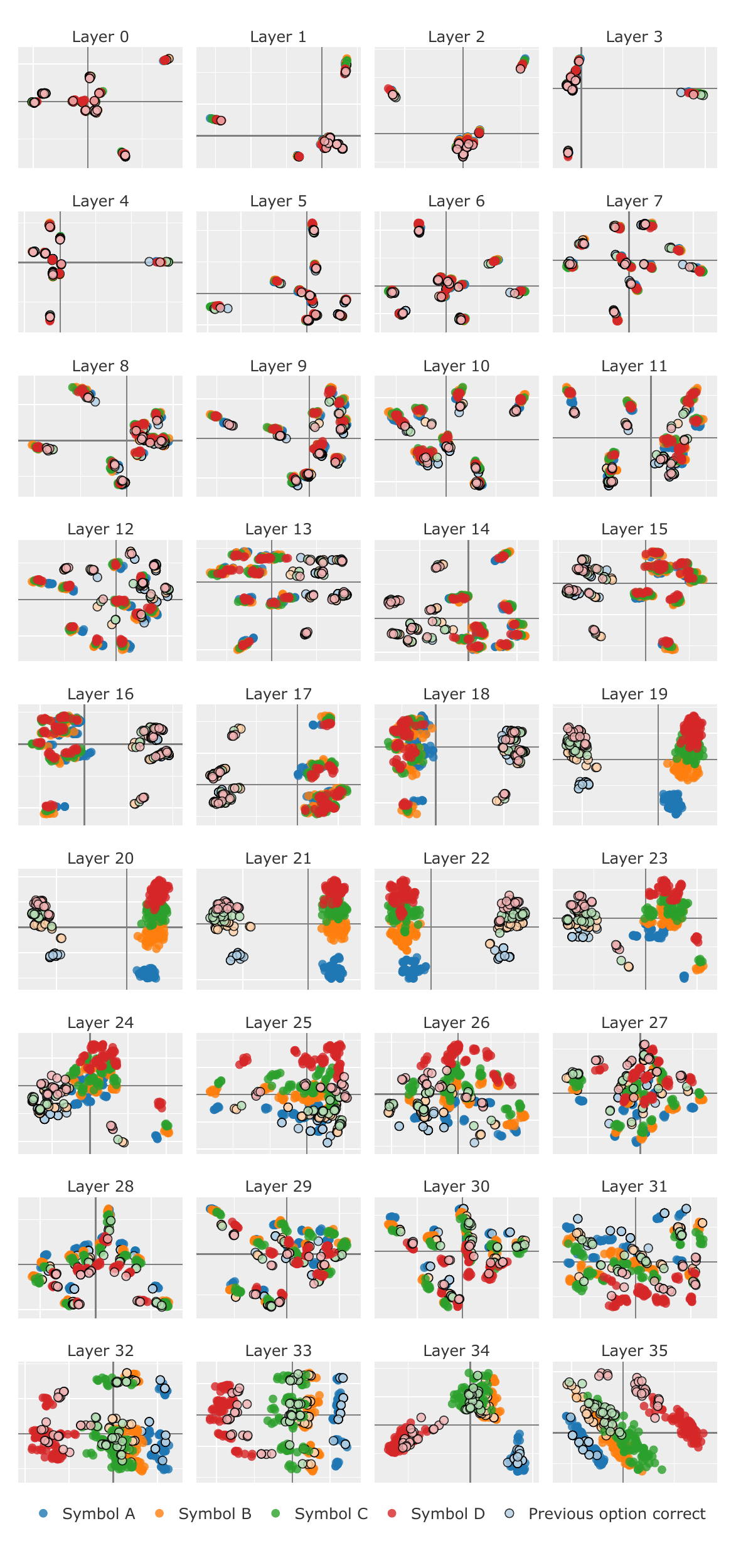}
    \caption{PCA for hidden activations extracted from \textbf{Qwen3-8b} at the \textbf{content token} positions.}
    \label{fig:appendix_pca_qwen3_content}
\end{figure}

\begin{figure}
    \centering
    \includegraphics[width=\linewidth]{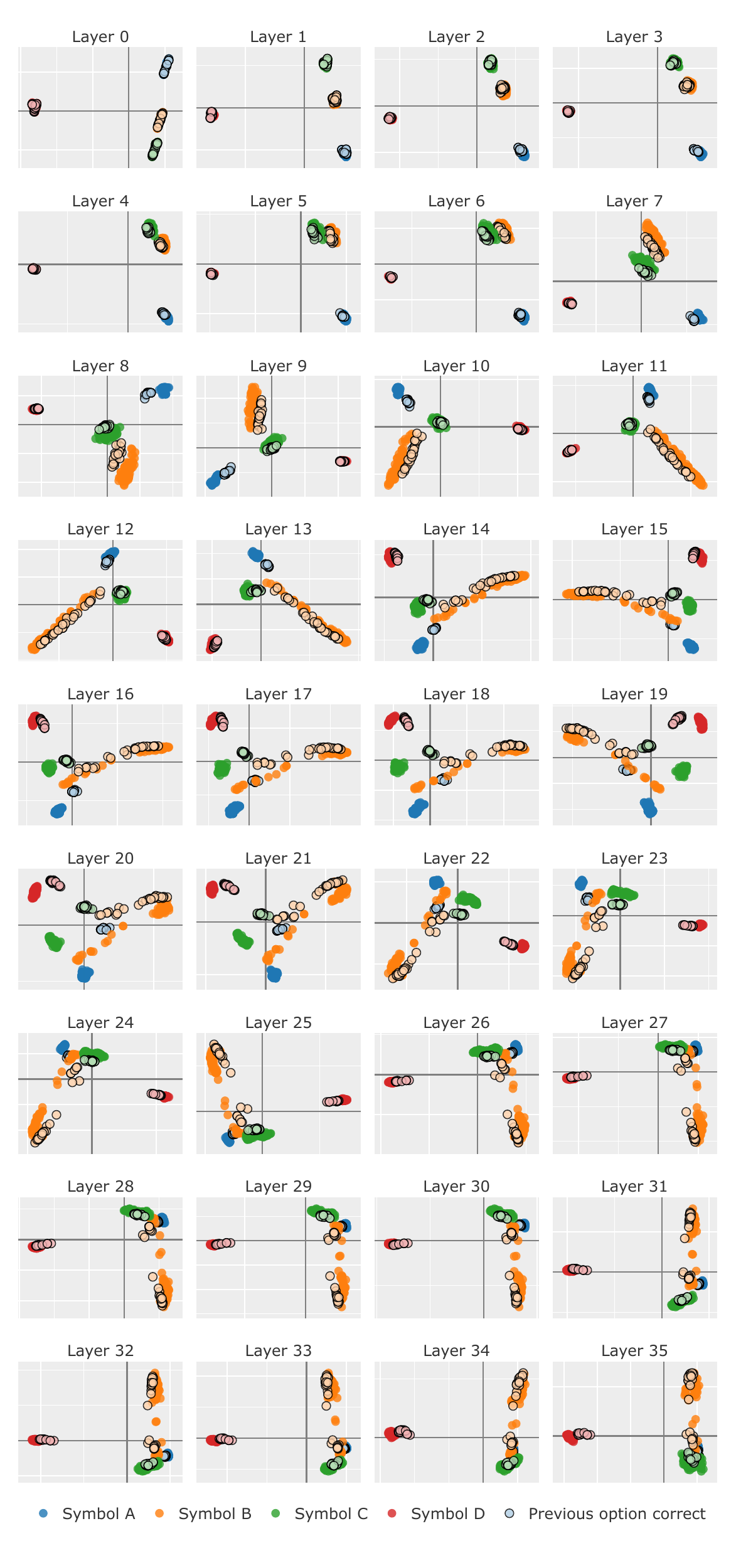}
    \caption{\textbf{Qwen3-8b} at the \textbf{newline tokens} directly after the content tokens.}
    \label{fig:appendix_pca_qwen3_newline}
\end{figure}

The following plots are the full versions of the layer-specific plots displayed in Figure~\ref{fig:pca_content_newline_subsets}.
Figures \ref{fig:appendix_pca_llama_content} and \ref{fig:appendix_pca_llama_newline} show the PCA plots for \textbf{Llama-3.2-3B-Instruct}, with hidden activations extracted at the \textbf{content tokens} and \textbf{newline} positions, respectively. Figures \ref{fig:appendix_pca_qwen3_content} (\textbf{content}) and \ref{fig:appendix_pca_qwen3_newline} (\textbf{newline}) do the same for \textbf{Qwen3-8b}.

\subsection{PCA with ARC-Easy}
\begin{figure}
    \centering
    \includegraphics[width=\linewidth]{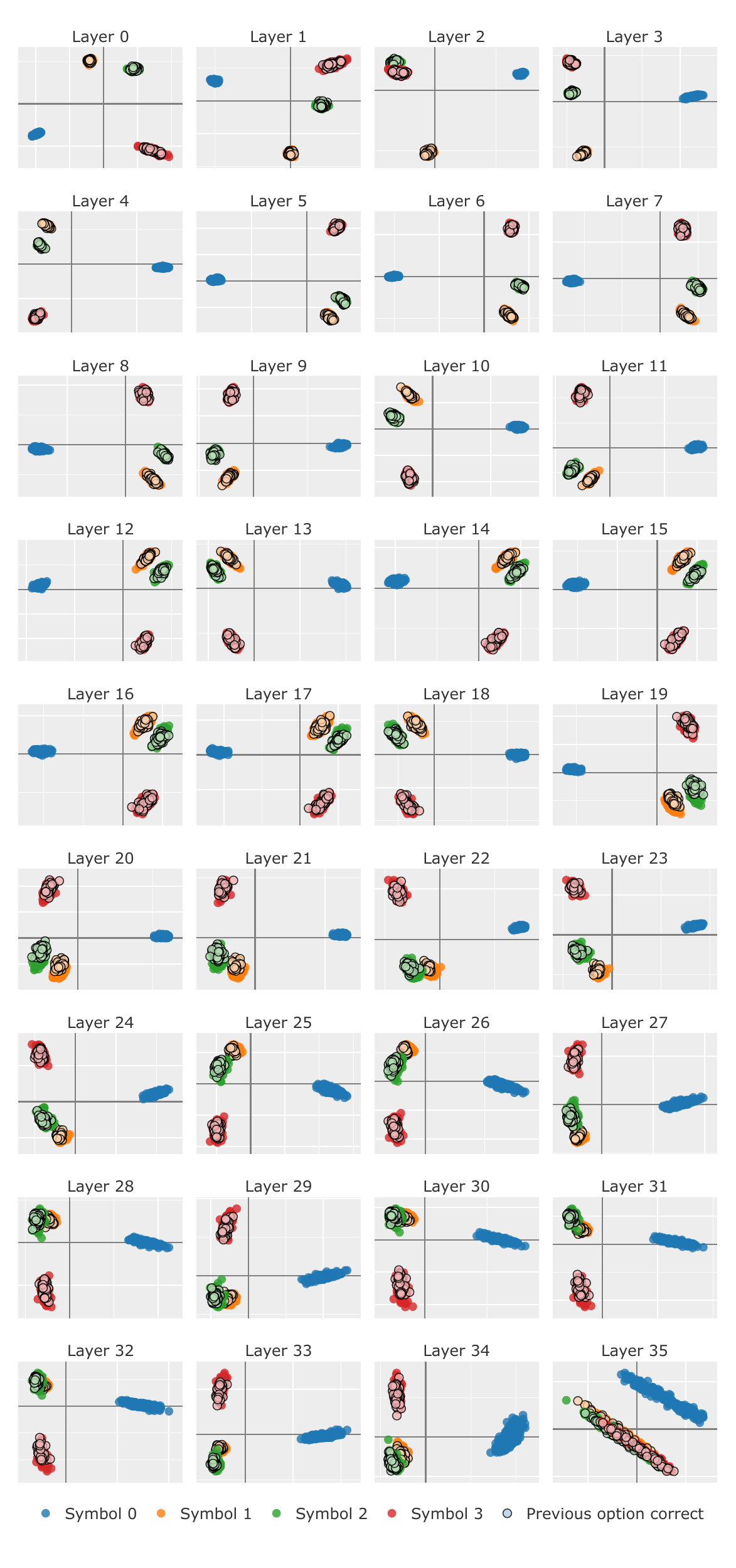}
    \caption{\textbf{Qwen3-8b} at the \textbf{symbol tokens} of 200 randomly selected ARC-Easy questions. Given that ARC-Easy is less structured than Colors, it is to be expected that patterns are not as easily visible in PCA space. However, do note that layers such as Layer 20 do hint at separability between correct and incorrect options.}
    \label{fig:appendix_pca_qwen3_arc}
\end{figure}

\begin{figure}
    \centering
    \includegraphics[width=\linewidth]{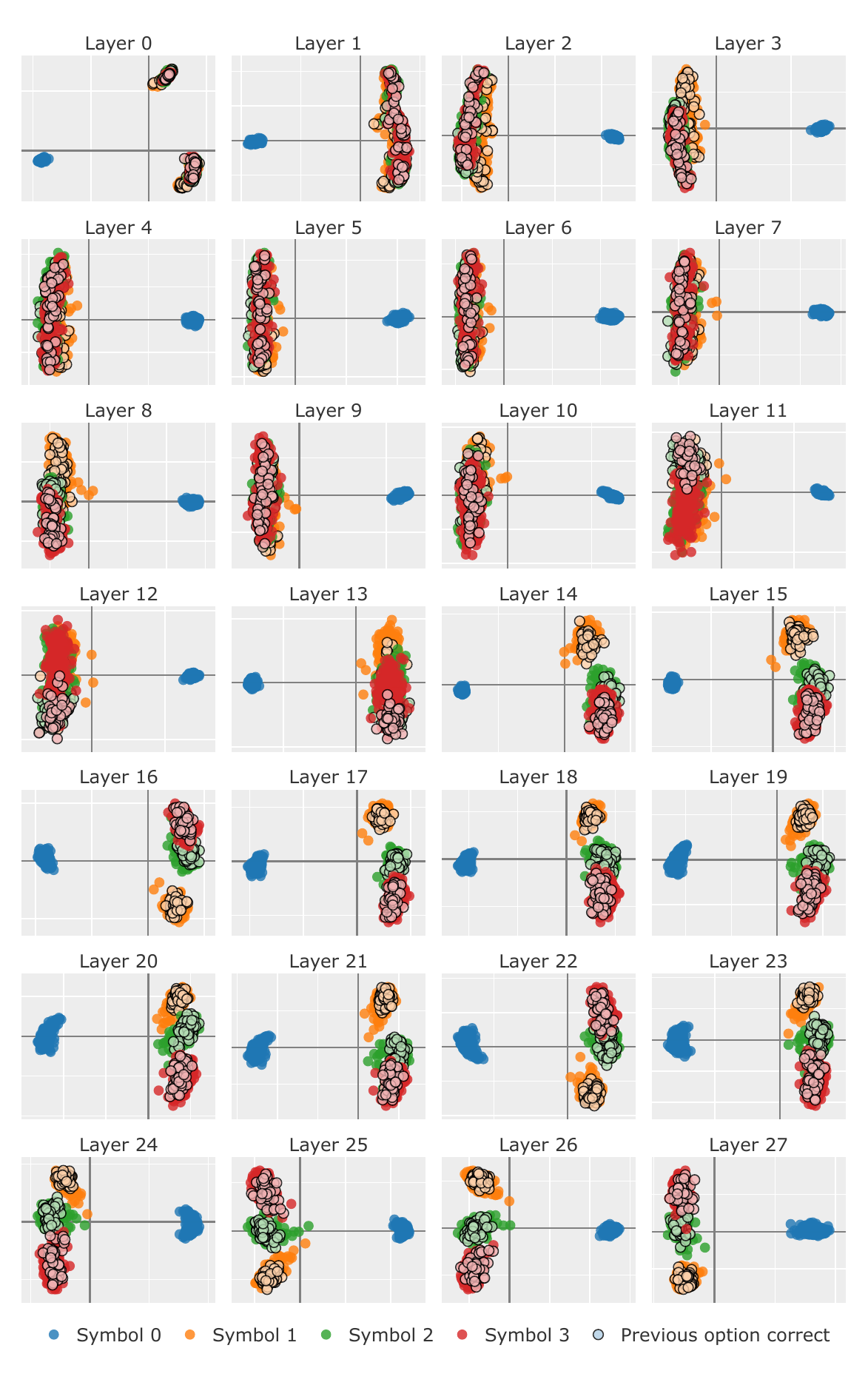}
    \caption{\textbf{Llama-3.2-3B-Instruct} at the \textbf{symbol tokens} of 200 randomly selected ARC-Easy questions. Given that ARC-Easy is less structured than Colors, it is to be expected that patterns are not as easily visible in PCA space. However, do note that layers such as Layer 12 do hint at separability between correct and incorrect options.}
    \label{fig:appendix_pca_llama_arc}
\end{figure}

Figures~\ref{fig:appendix_pca_qwen3_arc} and \ref{fig:appendix_pca_llama_arc} show results for PCA for the two models on ARC-Easy data. Given that ARC-Easy is less structured than Colors, it is to be expected that patterns are not as easily visible in PCA space. However, certain layers show separability between correct and incorrect options.

\label{appendix:pca}









\end{document}